\def\year{2020}\relax
\documentclass[letterpaper]{article} 

\usepackage{xcolor,colortbl}

\usepackage{aaai20}  
\usepackage{times}  
\usepackage{helvet} 
\usepackage{courier}  
\usepackage[hyphens]{url}  
\usepackage{graphicx} 
\usepackage{graphicx}  
\usepackage{todonotes}

\colorlet{newColor}{green!5!orange!95!}

\usepackage{enumitem}
\usepackage{multirow}
\usepackage{hhline}
\usepackage{booktabs}
\usepackage{subcaption}

\usepackage{makecell}
\usepackage{graphicx}

\usepackage{url}

\usepackage{amsmath}
\usepackage{algorithm}
\usepackage[noend]{algpseudocode}

\title{Practical Solutions for Machine Learning Safety in Autonomous Vehicles}

\author{Sina Mohseni,\textsuperscript{\rm 1,2}
Mandar Pitale,\textsuperscript{\rm 2}
Vasu Singh, \textsuperscript{\rm 3}
Zhangyang Wang \textsuperscript{\rm 1}\\
\textsuperscript{\rm 1} Texas A\&M University, College Station, TX \\
\textsuperscript{\rm 2} NVIDIA, Santa Clara, CA\\
\textsuperscript{\rm 3} NVIDIA, Munich, Germany\\
\{sina.mohseni,atlaswang\}@tamu.edu, \{mpitale,vasus\}@nvidia.com
}

\begin{document}

\maketitle
\begin{abstract}

Autonomous vehicles rely on machine learning to solve challenging tasks in perception
and motion planning. 
However, automotive software safety standards have not fully evolved to address the challenges of machine learning safety such as interpretability, verification, and performance limitations.
In this paper, we review and organize practical \textit{machine learning safety} techniques that can complement \textit{engineering safety} for machine learning based software in autonomous vehicles. 
Our organization maps safety strategies to state-of-the-art machine learning techniques in
order to enhance dependability and safety of machine learning algorithms. 
We also discuss security limitations and user experience aspects of machine learning components in autonomous vehicles.

\end{abstract}
\section{Introduction}

Advances in machine learning (ML) have been one of the biggest innovations of the last decade. Nowadays, ML models  are used extensively in different industrial fields like autonomous vehicles, medical diagnosis, and robotics to perform various tasks such as speech recognition, object detection, and motion planning.
Among different ML models, Deep Neural Networks (DNNs)~\cite{lecun2015deep} are well-known and widely used for their powerful representation learning in high-dimensional data.
For instance, in the field of autonomous driving, various DNN object detection and image segmentation algorithms have been used as perception units to process camera (e.g., PilotNet~\cite{bojarski2016end}, Fast RCNN~\cite{wang2017fast}) and Lidar (e.g., VoxelNet~\cite{zhou2018voxelnet}) data.

The development of safety critical systems relies on stringent safety methodologies, designs, and analyses to prevent hazards at the time of failure.
In the automotive field, ISO26262 and ISO/PAS 21448 are two main safety standards used to address safety of electrical and electronic components.
These standards mandate methodologies for system, hardware, and software development.
Specifically for software development, the process ensures traceability across requirements, architectural and unit design, code, and verification.
In cases of autonomous software with high complexity, iterative hazard analysis and risk assessment of autonomous software is required to formally represent the operational design domain.

On the other hand, ML models have numerous inherent safety drawbacks including accuracy on their training set and robustness in operational domain in open-world limitations.
For example, ML models are fragile to domain shift~\cite{ganin2014unsupervised}, data corruption and natural perturbations~\cite{hendrycks2019benchmarking}.
Also, prediction probability scores in DNNs do not provide a true representation of model uncertainty.
Moreover, from a security perspective, it has been shown that DNNs are susceptible to adversarial attacks that make small perturbations to the input sample (indistinguishable by human eye) but can fool a DNN~\cite{goodfellow2014explaining}.
Due to the lack of verification techniques for DNNs, validation of ML models often relies on simple accuracy measures on different large test sets to cover the targeted operation design domain.
While an important metric to gauge the success of the algorithm, it is definitely insufficient to measure performance in safety-critical applications as the real-world examples may differ from the test set.

With the realization that ML models will be increasingly used in safety critical systems, we need to investigate gaps these models expose in existing engineering safety standards.
Several examples of these gaps have been discussed in machine learning safety including interpretability and traceability into code, formal verification, and design specification~\cite{salay2017analysis}.

In this paper, we review challenges and opportunities in algorithmic techniques for 
ML safety to complement existing software safety standards for autonomous vehicles. 
Section~2 briefly reviews the two main automotive industry safety standards and identifies their five fundamental gaps with machine learning algorithms. 
Section~3 describes practical algorithmic safety techniques by surveying machine learning research on 1) error detectors and 2) model robustness. 
We briefly present three of our own implementations for safety-critical applications.
Section~4 discusses open challenges, directions for future work, and concludes the paper.

\section{Background}

Industrial safety broadly refers to the management of all operations and events within an industry in order to protect its employees and users by minimizing hazards, risks, and accidents.
Given the importance of correct operation of Electric and Electronic (E/E) components, the IEC 61508 is a basic functional safety standard developed for electrical and electronic safety related systems with two fundamental principles:
(a) Safety life cycle: an engineering process based on best practices to discover and eliminate design errors,
(b) Failure analysis: a probabilistic approach to account for the safety impact of system failures.
IEC 61508 has several derivations specific to different industries. 
For example, to achieve safety in the automotive industry, engineers are required to follow the ISO 26262 standard to minimize safety risks due to E/E faults to an acceptable level.

In this section, we briefly review the two main automotive safety standards followed by an organized list of fundamental ML limitations to meet safety requirements.
We note that both automotive standards mandate a meticulous analysis of hazards and risks, followed by a detailed development process focusing on system requirements, documented architecture and design, well structured code, and a thorough verification strategy for unit, integration, and system-level testing.

\subsection{ISO 26262 Standard}

\begin{figure}
    \centering
    \includegraphics[width=.9\columnwidth]{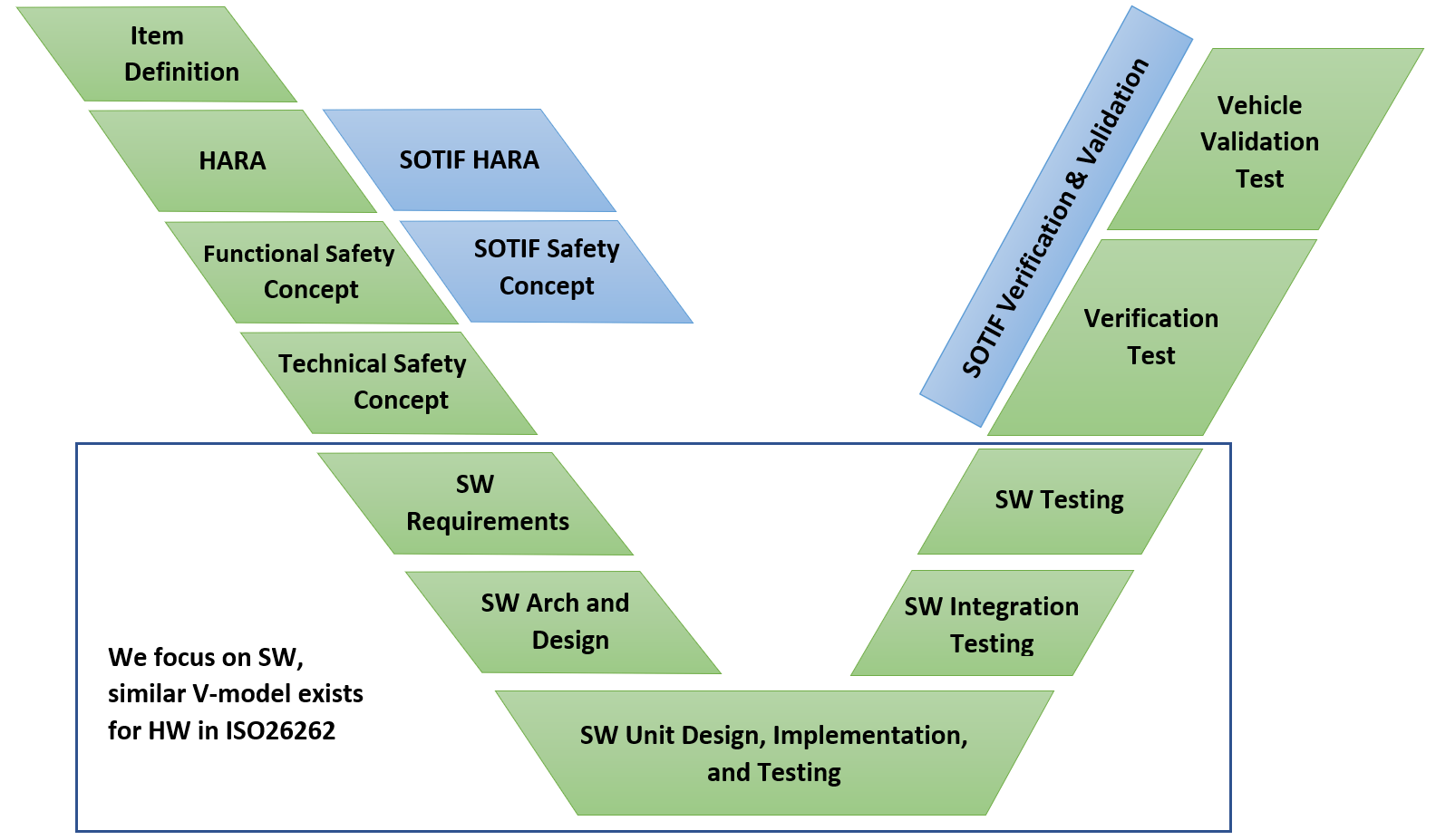}
    \caption{Comparison between the V-model in ISO 26262-6 (in Green) and ISO/PAS 21448 (in Blue) for software (SW) safety process. 
    }
    \label{fig:both_standards}
\end{figure}

ISO 26262 or the automotive E/E functional safety standard defines vehicle safety as the absence of unreasonable risk due to malfunctioning E/E components.
It requires a Hazard Analysis and Risk Assessment (HARA) to determine vehicle level hazards.
The potential hazards and risks guide the safety engineers towards safety goals that are then used to create functional safety requirements. 
These safety requirements then guide the system development process which is then decomposed into hardware and software development processes. 
Figure~\ref{fig:both_standards} presents an overview of this standard.
Our focus in this paper is Part 6 of ISO 26262 that defines the V-model for the software development process. 

The objective of the V-Model in Figure~\ref{fig:both_standards} is to ensure that software safety requirements are sufficiently fulfilled by the software architectural design and sufficiently tested by the software verification tests.
Similarly, the software architectural design is verified and the software integration tests prove that the interactions between the architectural entities including the static and dynamic aspects are tested. 
At the lowest level of the V-model, the unit design specifies the design details of each unit (that is identified during software architectural design) such as input, output error handling, behavior of the unit so that it can be coded.
Finally, the unit tests ensure that the unit is tested so that its requirements and design aspects are met and sufficient structural coverage of the unit is met.

ISO 26262 also identifies the methods for fault detection and avoidance to minimize the risk to an acceptable level. 
However, when applied to autonomous driving, ISO 26262 has several limitations. 
It cannot account for faults occurring due to the inability of the component to comprehend the environment - for example due to performance limitations or robustness issues - or foreseeable misuse of the system.

\subsection{ISO/PAS 21448 Standard}

ISO/PAS 21448 or SOTIF (Safety Of The Intended Functionality) standard describes an iterative development process consisting of design specification, development, and verification and validation phases.
SOTIF standard recognizes performance limitations of software (including ML components) and expects that the scenarios/inputs that belong to unsafe-unknown (e.g., samples out of training distribution) and unsafe-known (e.g., samples out of operational design domain) situations shall be reduced to the extent such that the residual error risk is acceptable.

The SOTIF process augments the ISO 26262 with a SOTIF specific HARA and safety concept. 
The hazard analysis consists of identifying the hazards due to inadequate performance functionality, insufficient situational awareness, reasonably foreseeable misuse or shortcomings of the human machine interface. 
In comparison, the hazard and risk analysis of ISO 26262 is limited to hazards due to E/E failures. 
If the SOTIF analysis results in risk that is higher than acceptable, functional modifications are performed to reduce SOTIF risks. 
A verification and validation strategy is then developed to argue that the residual risk is below the acceptable level.

\subsection{Safety Gaps for ML Components}

In recent years, there has been an increasing attention in identifying safety limitations of ML models.
For example, Varshney et al.~\cite{varshney2016engineering} discussed the definition of safety for ML models and compared it with four main \textit{engineering safety} strategies (1) inherently safe design, (2) safety reserve, (3) safe fail, and (4) procedural safeguards) in industry.
In a review of automotive software safety methods, Salay et al.~\cite{salay2017analysis} presented an analysis of ISO-26262 part-6 methods with respect to safety of ML models.
Their assessment of applicability of the software safety methods on machine learning algorithms (as software unit design) shows about 40\% of software safety methods do not apply to ML models.

Safety and robustness of ML models problem has also been in the interest of ML scientists~\cite{hernandez2019surveying}.
Amodei et al.~\cite{amodei2016concrete} presented five concrete research problems that could result in unintended and unsafe behavior of real-world AI systems.
They focus and characterize safety problems of AI into defining and evaluating the objective functions. 
Also, Ortega et al.\cite{ortega2018threepilar} introduced three areas of technical AI safety as specification (in design and emergent), robustness (for error prevention and recovery), and assurance (for monitoring and enforcement). 

We categorize these open challenges in the terminology of automotive software safety standards and briefly review representative research (see first row of Table~\ref{tab:table-2}) to close these safety gaps.

\begin{itemize}
    \item \emph{Design Specification.}
    Documenting and reviewing the software specification is a crucial step in functional safety, however, the design specification of ML models is generally not adequate, as
    the models learn the patterns in data to discriminate or generate  their distribution for new unseen input.
    Therefore, machine learning algorithms learn the target classes through their training data (and regularization constraints) and not using a formal specification.
    The lack of specifiability could cause mismatch between ``designer objectives'' and ``what the model actually learned'' which could result in unintended functionality of the system.
    This data-driven optimization of variables to train machine learning models makes it impossible to define and pose specific safety constraints.
    To this end, Seshia et al.~\cite{seshia2018formal} surveyed the landscape of formal specification for DNNs to lay an initial foundation for formalizing and reasoning about properties of DNNs.
    Another practical way to manage this design specification problem is to break machine learning components into smaller algorithms (with smaller tasks) to work in hierarchical structures.
    Related to this, Dreossi et al.~\cite{dreossi2019verifai} presented VerifAI toolkit for formal design and analysis of AI-based systems.

\item \emph{Implementation Transparency.}
    ISO26262 requires traceability from requirements to design. 
    However, advanced ML models trained on high dimensional data are not transparent.
    The very large number of variables in the models makes them incomprehensible or so-called black-box for design review and inspection. 
    In order to achieve traceability, significant research has been performed on interpretability methods for DNN to provide instance explanations of model prediction and DNN intermediate feature layers~\cite{zeiler2014visualizing}.
    In autonomous vehicles application, using the VisualBackProp technique~\cite{bojarski2018visualbackprop} shows that a DNN algorithm trained to control a steering wheel would in fact learn patterns of lane, road edges, and parked vehicles to execute the targeted task.
    However, the completeness of interpretability methods to grant traceability is not proven yet~\cite{adebayo2018sanity} and in practice interpretability techniques are mainly used by designers to improve network structure and training process rather than support a safety assessment. 

\item \emph{Testing and Verification.}
    Significant verification of work products are required for unit testing to meet the ISO 26262 standard. 
    For example, coding guidelines for software safety enforce that there is no dead or unreachable code. Depending upon the safety integrity level, complete statement, branch coverage, or modified condition and decision coverage are required to confirm adequacy of the unit tests. 
    Coming to DNNs, formally verifying their correctness is challenging (provably NP-hard~\cite{seshia2016towards,seshia2016towards}) due
    to the high dimensionality of the data.
    Therefore, reaching a complete validation and testing bounded to the operational design domain is difficult.
    As a result, researchers proposed new techniques such as searching for unknown-unknowns~\cite{bansal2018coverage} and predictor-verifier training~\cite{dvijotham2018training}.
    Other techniques including neuron coverage and fuzz testing\cite{wang2018efficient} in neural networks covers these aspects.
    Note that formal verification of shallow and linear models for low dimensional sensor data does not have the challenges of DNN verification.

    \item \emph{Performance and Robustness.}
    SOTIF standard treats the ML models as a black box and suggests using methods to improve model performance and robustness.
    However, improving model performance and robustness in itself is a very challenging task.
    In learning problems, typically training a model ends with an error rate (due to false positive and false negative predictions) on the training set.
    \textit{Empirical error} is prediction error rate of the learner function  on its target distribution.
    \textit{Generalization error} refers to the gap between model's empirical error on its training and test sets. 
    On top of these, \textit{operational error} is referred to model's error rate on open-world deployment that could be higher than test set error rate.
    Domain generalization refers to ability of model in learning generalizable data representations for open-world tasks.
    We will review more details and machine learning techniques to improve model robustness in Section~3.
    
    \item \emph{Run-time Monitoring Function.}
    SOTIF and ISO 26262 standards suggest run-time monitoring functions as software error detection solutions.
    Monitoring functions in classical software are based on a rule-set to detect cases like transient hardware error, software crash, and exit from operation design domain.
    However, designing monitoring functions to predict ML failure (e.g., false positive and false negative error) is different in nature. 
    ML models generate prediction probability for input instances but research shows prediction probability does not guarantee failure  prediction~\cite{hein2019relu}.
    In fact, DNN and many other ML models could generate incorrect outputs with high confidence in cases of distribution shift and adversarial attacks.
    We will review more details of error detection techniques for ML models in Section 3.
    
\end{itemize}  
\section{Techniques for ML Safety}

We introduce algorithmic techniques for ML safety to maintain dependability of machine learning algorithms for safe execution in open-world tasks. 
The techniques we review in this section are meant to complement classic engineering strategies in software safety.
We also connect ML safety techniques with appropriate engineering safety strategies (see Table~\ref{tab:table-2}) to help machine learning scientists and safety engineers find a common ground in the new topic of autonomous vehicle safety.
We follow Varshney's~\cite{varshney2016engineering} four strategies to achieve machine learning safety to map AI Safety techniques with engineering strategies.

Considering the research debt left to assure different safety aspects of AI (see Section 2), we believe we are far from reaching \textit{(1) Inherently Safe} AI.
Therefore, we focus on practical machine learning solutions for the two following safety strategies:

\begin{itemize}
    \item \emph{(2) Safe Fail} refers to strategies to keep the vehicle on road in a safe state at the time of failure.
    This strategy can mitigate hazards at the time of fault by use of monitoring functions and graceful degradation plans such as notifying the driver to take vehicle's control.
    We propose using run-time \textit{error detection} techniques to detect erroneous output of machine learning algorithms (e.g. misclassification and misdetection) for the vehicle on the road.

    \item \emph{(3) Safety Margins} in the context of machine learning is described as difference between model's performance on training set and operational performance in open-world. 
    We propose using \textit{model robustness} techniques to improve resilience and hence Safety Margin of machine learning components. 
\end{itemize}

We will also briefly review the importance of \textit{(4) Procedural Safeguards} for non-experts end-users of autonomous vehicles (i.e. driver and passengers) later in the discussion about future work.
To separate safety and security concerns, we consider external factors that intentionally exploit system vulnerabilities (e.g., sample manipulations in adversarial attack) as security concerns but not safety. 

\subsection{Monitoring Function}

Our first practical \textit{ML safety} solution leverages from the range of techniques for machine learning misclassification error detection to achieve \textit{Safe Fail} behavior.  
For example, when transient errors in hardware like sensors affect the functionality of the software like that for cruise control, an error detection unit (monitoring function) can detect the error and degrade the system by appropriate warnings and allowing the driver to take over.
Similarly, various run-time monitoring functions and error detectors could be designed for machine learning components to predict model failure and trigger appropriate warnings.
In the following we introduce three types of error detectors for machine learning and review their relations and limitations.
Note that although the three following groups of error detectors overlap, we separate detectors by their targeted error types.

\subsubsection{Uncertainty Estimation}
Uncertainty in probabilistic learners is an important factor to maintain fail-safety of the system. 
Even well trained and calibrated predictors that are robust to noise, corruption, and perturbations can benefit from uncertainty estimation to detect domain shift and out-of-distribution samples at run-time. 
Quantifying uncertainty can explain what a model does not know in terms of model confidence on its prediction (epistemic uncertainty or model uncertainty) and uncertainty for unknown samples (aleatoric uncertainty or data uncertainty).

Regarding importance of uncertainty methods for safety critical applications, McAllister et al.~\cite{mcallister2017concrete} proposed measuring uncertainties in ML models and propagating down in the decision-making pipeline as a key for safety of autonomous systems.
However, quantifying predictive uncertainty in DNNs is a challenging task.
Typically, DNN classification models generate normalised prediction scores which tend to be overconfident and regression DNN models do not give uncertainty representation in their output. 
Research in DNN presents solutions such as deep ensembles \cite{lakshminarayanan2017simple} and Monte Carlo dropout (MC-dropout)~\cite{gal2016dropout} to estimate prediction uncertainty.
Uncertainty estimation methods have been tested for various model error types~\cite{kendall2017uncertainties} including adversarial attack detection~\cite{smith2018understanding}.

Although uncertainty estimation methods offer potential effective solutions for DNN failure prediction -- in practice -- they carry significant computation cost and latency which is not ideal for run-time failure prediction.
For example, to design an error detector for a PilotNet algorithm, Michelmore et al.~\cite{michelmore2018evaluating} present an implementation of MC-dropout uncertainty estimation which needs 128 stochastic forward passes to estimate uncertainty of the model.
Therefore, in resource limited setups and for reasons like computation simplicity, researchers work on alternative error detection solutions which we review in the next two subsection.

\begin{figure}[t]
\centering
  \includegraphics[width=1.0\columnwidth]{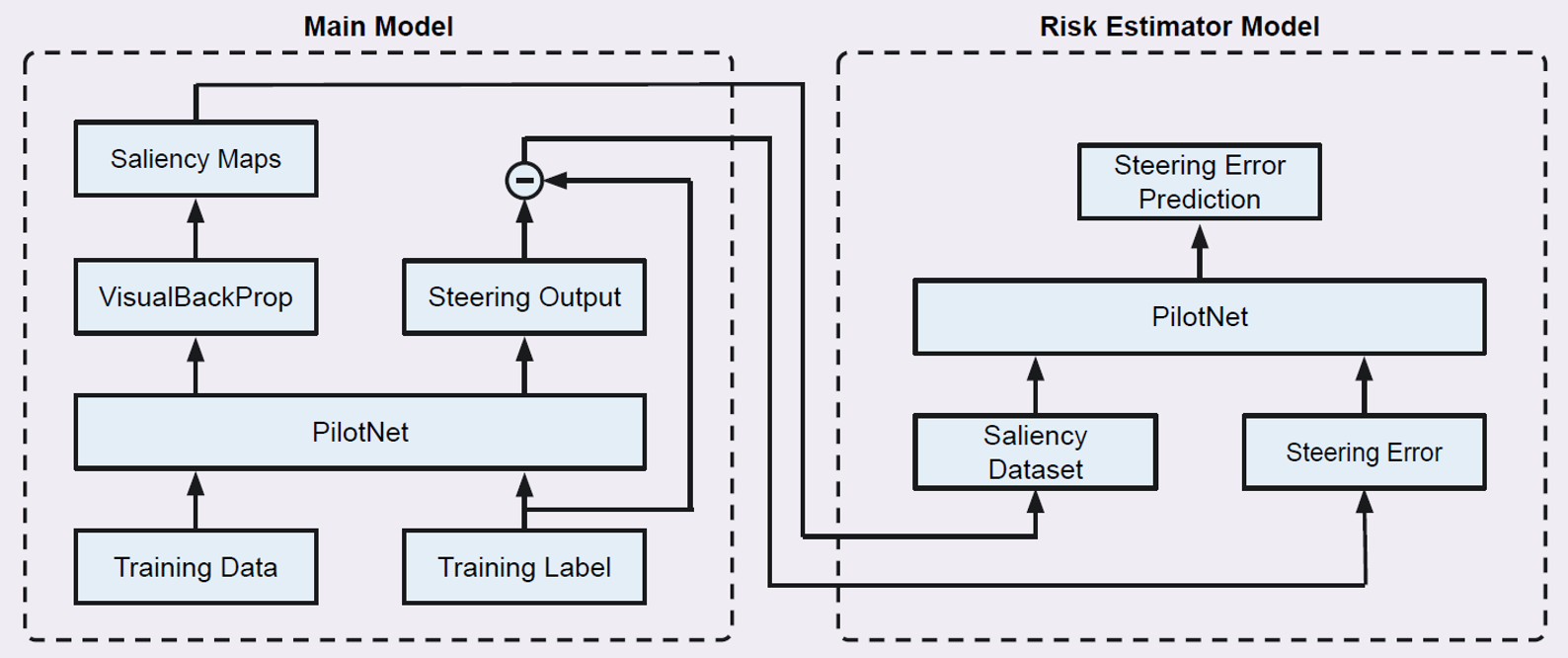}
  \caption{
  Model failure prediction based on saliency map for ConvNet regression models. We trained a student model as monitoring function for run-time failure predictor of a PilotNet model~\cite{mohseni2019predicting}.
  }
  \label{fig:student-teacher}
\end{figure}

\subsubsection{In-distribution Error Detectors}
Misclassification of in domain samples often happens due to weak representation learning. 
In recent years, advanced neural networks, regularization techniques, and large training datasets significantly improved DNNs representation learning and therefore model performance and robustness.
However, run-time prediction error detectors are still needed to maintain safety of the system in case of model failure.
Selective classification (also known as classification with reject option) is a technique to cautiously provide prediction for high-confidence samples and abstain when in doubt.
This method for confident prediction can significantly improve model performance at the cost of test coverage.
Geifman and El-Yaniv~\cite{geifman2017selective} present a simple and effective implementation of selective classification for DNNs.
They introduce a reject function which guarantees control over the true risk based on DNN softmax output.
They later introduce SelectiveNet~\cite{geifman2019selectivenet}, a three-headed network to jointly train classification and rejection functions on the normal domain.
Along similar lines, Guo et al.~\cite{guo2017calibration} presented temperature scaling as a post-processing calibration technique to adjust the model probability estimates being off due to over fitting.
In an application to autonomous vehicles, Hecker et al.\cite{hecker2018failure}  added and trained a failure head to the network in order to learn to predict the occurrence of model failures.

In our recent paper, we presented an error detector for regression models in autonomous vehicle applications~\cite{mohseni2019predicting}.
We proposed a new design which trains a student model (\textit{failure predictor}) for predicting the teacher model's (\textit{main model}) error at run-time. 
Figure~\ref{fig:student-teacher} shows how the student model learns teacher model's prediction loss on a validation set in order to predict its failures on the test set.
We also train the student model using the saliency maps from the main model to improve failure prediction performance.
We evaluate our failure predictor model based on prediction error and the driving safety gained by this system. 

\begin{figure}[t]
\centering
  \includegraphics[width=1.0\columnwidth]{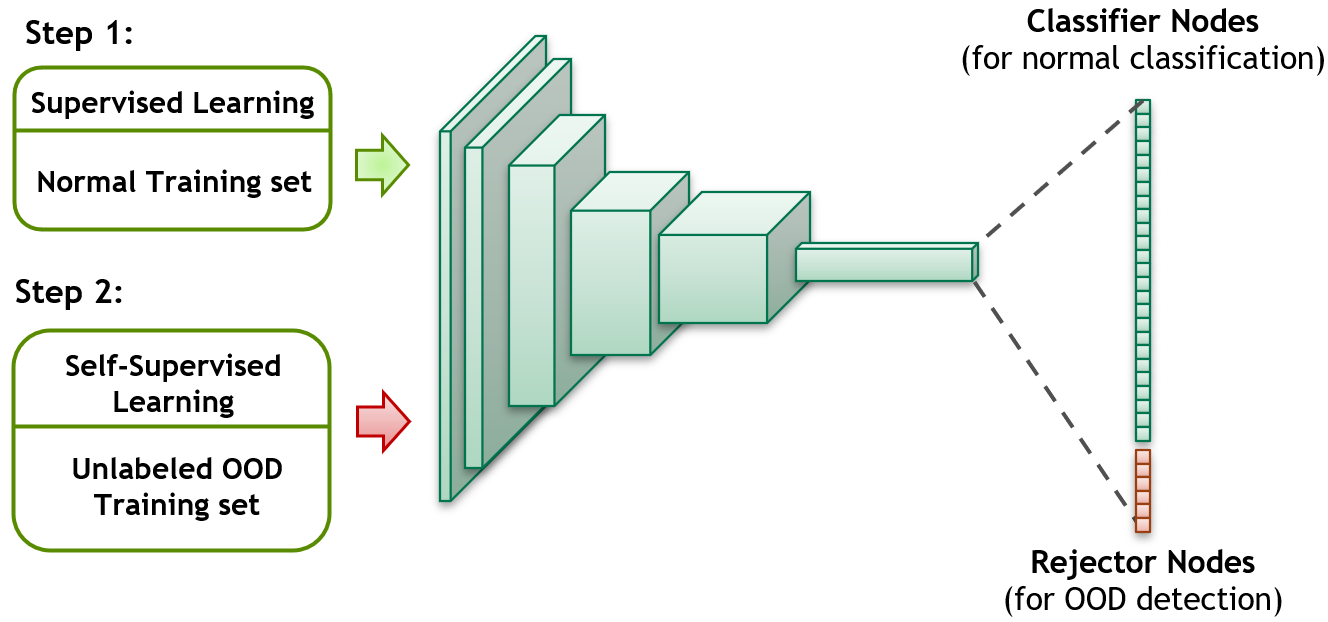} 
  \caption{
  Out-of-distribution sample detection base on prediction confidence.
  We proposed an out-of-distribution sample detector to train reject classes to learn outlier features in a self-supervised step. ~\cite{mohseni2020ood}
  }
  \label{fig:ood-figure}
\end{figure}

\subsubsection{Out-of-distribution Error Detectors}
Out-of-distribution (OOD) sample or outlier refers to inputs that are outside the normal training distribution. 
OOD error is referred to ML model misclassification error for OOD samples. 
Examples of OOD samples in autonomous vehicles include unique, unusual or unknown road signs, road marks, or rare object or scenario which either was not included in the training set or the model was not able to learn (e.g. due to class imbalance) during the training process.
OOD error is an inherent problem with ReLU family activation functions that they produce arbitrary high confidence as inputs get further from the training distribution~\cite{hein2019relu}.
However, various techniques known as OOD detector, novelty detector, and outlier detector have been proposed to detect OOD samples.
Examples of OOD detector techniques include revising network architecture for learning the prediction-confidence~\cite{devries2018learning}, employing ensemble of leaving-out classifiers~\cite{vyas2018out}, and self-supervised representation learning~\cite{golan2018deep} approaches for outlier detection.
On the other hand, a fast and low cost approach in OOD detection is to use class probabilities as a measure for OOD detection~\cite{hendrycks2016baseline}.
In this regard, new techniques have been proposed to calibrate DNN decision-boundaries for robust OOD detection ~\cite{lee2017training}.

In our recent work, we present a fast and memory efficient OOD error detection technique by embedding and training reject options into any DNN discriminative model with minimal architectural changes~\cite{mohseni2020ood}. This is shown in Figure~\ref{fig:ood-figure}.
Our fundamental idea is to exploit the high-level feature learning capacity of DNNs to jointly learn generalizable outlier features as well as in-distribution features for normal classification both in one network.
Figure~\ref{fig:ood-figure} shows how we train additional reject functions in the last layer of our neural network using a two steps of supervised (with labeled in-distribution training set) and self-supervised (with free unlabeled OOD natural samples) training. 
Our evaluation results show the proposed self-supervised learning of OOD features can very well generalize to reject other unseen distributions.

\begin{table*}[]
\centering
\caption{Table of practical machine learning techniques to improve the safety of ML algorithms. Left column presents engineering safety strategies and right column maps machine learning techniques with three representative research papers for each row.}
\label{tab:table-2}
\resizebox{\textwidth}{!}{%
\begin{tabular}{@{}cl@{}}
\toprule
\rowcolor[HTML]{FFFFFF} 
\textbf{Safety Strategy} & \textbf{Practical AI Safety Opportunities} \\ \midrule
 & Design Specification (~\cite{amodei2016concrete,leike2017ai,seshia2018formal}) \\ \cmidrule(l){2-2} 
 & Implementation Transparency (~\cite{bojarski2018visualbackprop,zeiler2014visualizing,adebayo2018sanity}) \\ \cmidrule(l){2-2} 
\multirow{-3}{*}{\begin{tabular}[c]{@{}c@{}}Inherently Safe \\ Design\end{tabular}} & Formal Verification (\cite{seshia2016towards,dvijotham2018training,wang2018efficient}) \\ \midrule
 & Uncertainty Estimation (\cite{lakshminarayanan2017simple,gal2016dropout}) \\ \cmidrule(l){2-2} 
 & In-distribution error detection (\cite{geifman2017selective,geifman2019selectivenet,guo2017calibration}) \\ \cmidrule(l){2-2} 
\multirow{-3}{*}{Safe Fail} & Out-of-distribution error detection (~\cite{mohseni2020ood,golan2018deep,vyas2018out}) \\ \midrule
 & Domain Generalization (\cite{zhang2019dada,ganin2014unsupervised,lee2019me}) \\
 \cmidrule(l){2-2} 
\multirow{-2}{*}{Safety Margin} & Perturbation and Corruptions Robustness (\cite{geirhos2018imagenet,huang2017multi,hendrycks2019using}) \\ \bottomrule
\end{tabular}%
}
\end{table*}

\subsection{Algorithm Robustness}
The second practical solution for \textit{ML safety} leverages robustness techniques to improve \textit{Safety Margins} of ML models in autonomous vehicles. 
Robustness techniques in ML research improve resiliency of the algorithms to unseen samples, natural corruptions and perturbations, adversarial example, and domain shifts.
ML literature present multiple techniques such as dataset augmentation, noise injection, and multi-task learning to regularize DNNs to learning generalizable features~\cite{goodfellow2016deep}. 
Other techniques, including transfer learning~\cite{hendrycks2019using}, have shown to improve model robustness by transferring universal representation from a pre-trained model to the new domain. 
Further, Zhang and LeCun~\cite{zhang2017universum} explored using unlabeled free data to regularize model training for its robustness.
In the following, we review two major safety-related types of machine learning robustness techniques for open-world tasks.
We will also briefly review robustness and detection techniques for adversarial examples in the discussion section.

\subsubsection{Robustness to Domain Shift}

Domain shift (also known as distribution shift and dataset shift) describes variations in the input data distribution in comparison to the training set.
Distribution shift reduces the operational performance compared to the test set performance by breaking the i.i.d assumption between training and testing data. 
In this regard, \textit{domain generalization} is a crucial aspect of machine learning algorithms for open-world applications such as autonomous vehicles which data is captured from uncontrolled and fast changing environment.
Domain generalization could be achieved from many different ways.
One approach is adversarial domain adaptation~\cite{ganin2014unsupervised} that leverages from massive amount of unlabeled data captured from the targeted domain.
For example, Zhang et al.~\cite{zhang2019dada} implemented learning-based approaches to synthesize foreground objects and background contexts for new training samples. 
Multi-task learning is another technique to improve model robustness by simultaneously learning two (or more) tasks.
For instance, Tang et al.~\cite{tang2019pamtri} presents a pose-aware multi-task vehicle re-identification techniques to overcome view-point dependency of the objects.
They created and used large-scale highly randomized synthetic dataset with automatically annotated vehicle attributes for training.
From a different approach, Lee et al.~\cite{lee2019me} propose using ensemble of models to capture and learn different poses and viewing angles of objects to improve overall robustness. 
Also, to improve robustness of object detection models to occlusions and deformations, Wang et al.~\cite{wang2017fast} used an adversarial network to generate hard positive examples. 

In a recent work~\cite{wu2019delving}, we proposed a new technique for improving model robustness to domain shift in unmanned aerial vehicle (UAV). 
We cast an object detection problem as a cross-domain object detection problem with multiple fine-grained domains.
We then train our object detection model to extract invariant features shared by many different ``non-ideal'' variations (e.g., weather conditions, camera angle, light conditions) of the target domains.
In order to do so, we add a nuisance disentangled feature transform block on the input and nuisance prediction branches (one for each non-ideal condition) in a modular fashion and jointly train the final network in an adversarial setting.
Our implementation on Faster-RCNN backbone~\cite{wang2017fast} shows superior results compared  to vanilla baseline on improving model robustness to weather, altitude, and view variations in UAV images.

\subsubsection{Robustness to Corruptions and Perturbations}

Natural data perturbation and corruption commonly exist in open-world settings.
Benchmarking DNN robustness to corruption and perturbations~\cite{hendrycks2019benchmarking} shows machine learning models exhibit unexpected prediction errors on simple perturbations.
Achieving model robustness to natural corruptions (e.g., due to camera lens, snow, rain, fog in image data) and perturbations (e.g., sensor transients error, electromagnetic interference on sensors) requires techniques to improve model robustness above their \textit{clean datasets}.
Previously, classical data augmentations were used to gain robustness to simple image variations like rotation and scaling~\cite{goodfellow2016deep}.
Other techniques such as using adaptive algorithm for choosing the augmentation transformations~\cite{fawzi2016adaptive} and random patch erasing~\cite{zhong2017random} also shown to be effective for both robustness and representation learning.
Recently, advanced augmentations such as style transfer~\cite{geirhos2018imagenet} have shown to improve model robustness to texture bias. 
Another line of research proposes larger networks~\cite{huang2017multi} to improve DNN robustness through multiscale and redundant feature learning.

Adversarial perturbation~\cite{goodfellow2014explaining}, on the other hand, are small but worst-case perturbations intentionally created by an attacker so that the perturbed sample results in the model misclassify the sample with high confidence.
We distinguish safety hazard due to natural perturbations from security hazard due to adversarial perturbations as the latter intentionally exploit system vulnerabilities to cause harm.
Security concerns related to adversarial perturbation are briefly mentioned in the discussion about future work.

\section{Conclusion and Future Work}

In this work, we presented a review and categorization of classical software safety methods with fundamental limitations in machine learning algorithms.
The impetus of this work was to leverage from both engineering safety strategies and state-of-the-art machine learning techniques to enhance dependability and safety of machine learning components in autonomous systems.
In this regard, maintaining the safety of autonomous vehicles requires a multidisciplinary effort across multiple fields including human-computer interactions, machine learning, software engineering, hardware engineering~\cite{koopman2017autonomous}.
We briefly review and discuss other dimensions of safety and security of ML-infused systems that could be benefit from research communities attention.

\noindent \textbf{Security Risks of Adversarial Attacks: }
\label{sec:adversarial_attack}
An adversarial example is a clean image delicately perturbed (by an adversary) with a small distortion so that it visually looks identical to the original clean image but is misclassified by the ML model.
Despite its popularity, adversarial attacks are not considered primarily a safety concern rather a security limitation~\cite{carlini2017adversarial}.
Two main defense approaches for adversarial attacks are~\textit{detection} and \textit{robustness}.
For example, Smith and Gal~\cite{smith2018understanding} present a case of MC-drop out uncertainty estimation technique for detecting adversarial examples.
Further, to improve model robustness and resiliency to adversary perturbation Papernot et al. proposed an effective defense based on model distillation~\cite{papernot2016distillation}. 
Still, the problem of adversarial examples is an unsolved problem when considering that attackers always react to current defense techniques by designing stronger attacks to undermine security of machine learning components~\cite{carlini2017adversarial}.

\noindent \textbf{Procedural Safeguards for ML Safety: }
Beyond functional safety of the system, procedural safeguards help operators and product end-users (e.g. driver in autonomous vehicle) avoid unintentional misuse of system due to lack of instructions and unawareness~\cite{varshney2016engineering}.
User experience (UX) design and algorithmic transparency are two approaches to improve safety of the operation in autonomous vehicles.
In such cases, end-users can benefit from explainable UX designs which provides useful and comprehensible information about model reasoning~\cite{gunning2017explainable} and prediction uncertainty~\cite{michelmore2018evaluating}.
For example, UX design could leverage real-time visualization of model uncertainty for vehicle detection and path planning to help improve driver's understanding of vehicle safety on the road.
\newline

In our future work, we plan to review recent procedural safeguards designs and studies for autonomous vehicle applications. 
Including key techniques and factors in calibrating user trust in human-AI systems. 

\bibliographystyle{aaai}
\bibliography{Bibliography}

\begin{thebibliography}{}

\bibitem[\protect\citeauthoryear{Adebayo \bgroup et al\mbox.\egroup
  }{2018}]{adebayo2018sanity}
Adebayo, J.; Gilmer, J.; Muelly, M.; Goodfellow, I.; Hardt, M.; and Kim, B.
\newblock 2018.
\newblock Sanity checks for saliency maps.
\newblock In {\em NIPS}.

\bibitem[\protect\citeauthoryear{Amodei \bgroup et al\mbox.\egroup
  }{2016}]{amodei2016concrete}
Amodei, D.; Olah, C.; Steinhardt, J.; Christiano, P.; Schulman, J.; and
  Man{\'e}, D.
\newblock 2016.
\newblock Concrete problems in ai safety.
\newblock {\em arXiv preprint arXiv:1606.06565}.

\bibitem[\protect\citeauthoryear{Bansal and Weld}{2018}]{bansal2018coverage}
Bansal, G., and Weld, D.~S.
\newblock 2018.
\newblock A coverage-based utility model for identifying unknown unknowns.
\newblock In {\em Thirty-Second AAAI Conference on AI}.

\bibitem[\protect\citeauthoryear{Bojarski \bgroup et al\mbox.\egroup
  }{2016}]{bojarski2016end}
Bojarski, M.; Del~Testa, D.; Dworakowski, D.; Firner, B.; Flepp, B.; Goyal, P.;
  Jackel, L.~D.; Monfort, M.; Muller, U.; Zhang, J.; et~al.
\newblock 2016.
\newblock End to end learning for self-driving cars.
\newblock {\em arXiv preprint arXiv:1604.07316}.

\bibitem[\protect\citeauthoryear{Bojarski \bgroup et al\mbox.\egroup
  }{2018}]{bojarski2018visualbackprop}
Bojarski, M.; Choromanska, A.; Choromanski, K.; Firner, B.; Ackel, L.~J.;
  Muller, U.; Yeres, P.; and Zieba, K.
\newblock 2018.
\newblock Visualbackprop: Efficient visualization of cnns for autonomous
  driving.
\newblock In {\em 2018 IEEE International Conference on Robotics and Automation
  (ICRA)},  1--8.
\newblock IEEE.

\bibitem[\protect\citeauthoryear{Carlini and
  Wagner}{2017}]{carlini2017adversarial}
Carlini, N., and Wagner, D.
\newblock 2017.
\newblock Adversarial examples are not easily detected: Bypassing ten detection
  methods.
\newblock In {\em Proceedings of the 10th ACM Workshop on Artificial
  Intelligence and Security},  3--14.
\newblock ACM.

\bibitem[\protect\citeauthoryear{DeVries and
  Taylor}{2018}]{devries2018learning}
DeVries, T., and Taylor, G.~W.
\newblock 2018.
\newblock Learning confidence for out-of-distribution detection in neural
  networks.
\newblock {\em ICLR}.

\bibitem[\protect\citeauthoryear{Dreossi \bgroup et al\mbox.\egroup
  }{2019}]{dreossi2019verifai}
Dreossi, T.; Fremont, D.~J.; Ghosh, S.; Kim, E.; Ravanbakhsh, H.;
  Vazquez-Chanlatte, M.; and Seshia, S.~A.
\newblock 2019.
\newblock Verifai: A toolkit for the formal design and analysis of artificial
  intelligence-based systems.
\newblock In {\em International Conference on Computer Aided Verification},
  432--442.
\newblock Springer.

\bibitem[\protect\citeauthoryear{Dvijotham \bgroup et al\mbox.\egroup
  }{2018}]{dvijotham2018training}
Dvijotham, K.; Gowal, S.; Stanforth, R.; Arandjelovic, R.; O'Donoghue, B.;
  Uesato, J.; and Kohli, P.
\newblock 2018.
\newblock Training verified learners with learned verifiers.
\newblock {\em arXiv preprint arXiv:1805.10265}.

\bibitem[\protect\citeauthoryear{Fawzi \bgroup et al\mbox.\egroup
  }{2016}]{fawzi2016adaptive}
Fawzi, A.; Samulowitz, H.; Turaga, D.; and Frossard, P.
\newblock 2016.
\newblock Adaptive data augmentation for image classification.
\newblock In {\em 2016 IEEE International Conference on Image Processing
  (ICIP)},  3688--3692.
\newblock Ieee.

\bibitem[\protect\citeauthoryear{Gal and Ghahramani}{2016}]{gal2016dropout}
Gal, Y., and Ghahramani, Z.
\newblock 2016.
\newblock Dropout as a bayesian approximation: Representing model uncertainty
  in deep learning.
\newblock In {\em ICML},  1050--1059.

\bibitem[\protect\citeauthoryear{Ganin and
  Lempitsky}{2014}]{ganin2014unsupervised}
Ganin, Y., and Lempitsky, V.
\newblock 2014.
\newblock Unsupervised domain adaptation by backpropagation.
\newblock {\em arXiv preprint arXiv:1409.7495}.

\bibitem[\protect\citeauthoryear{Geifman and
  El-Yaniv}{2017}]{geifman2017selective}
Geifman, Y., and El-Yaniv, R.
\newblock 2017.
\newblock Selective classification for deep neural networks.
\newblock In {\em NIPS},  4878--4887.

\bibitem[\protect\citeauthoryear{Geifman and
  El-Yaniv}{2019}]{geifman2019selectivenet}
Geifman, Y., and El-Yaniv, R.
\newblock 2019.
\newblock Selectivenet: A deep neural network with an integrated reject option.
\newblock {\em arXiv preprint arXiv:1901.09192}.

\bibitem[\protect\citeauthoryear{Geirhos \bgroup et al\mbox.\egroup
  }{2019}]{geirhos2018imagenet}
Geirhos, R.; Rubisch, P.; Michaelis, C.; Bethge, M.; Wichmann, F.~A.; and
  Brendel, W.
\newblock 2019.
\newblock Imagenet-trained cnns are biased towards texture; increasing shape
  bias improves accuracy and robustness.
\newblock {\em ICLR}.

\bibitem[\protect\citeauthoryear{Golan and El-Yaniv}{2018}]{golan2018deep}
Golan, I., and El-Yaniv, R.
\newblock 2018.
\newblock Deep anomaly detection using geometric transformations.
\newblock In {\em NIPS},  9758--9769.

\bibitem[\protect\citeauthoryear{Goodfellow, Bengio, and
  Courville}{2016}]{goodfellow2016deep}
Goodfellow, I.; Bengio, Y.; and Courville, A.
\newblock 2016.
\newblock {\em Deep learning}.
\newblock MIT press.

\bibitem[\protect\citeauthoryear{Goodfellow, Shlens, and
  Szegedy}{2014}]{goodfellow2014explaining}
Goodfellow, I.~J.; Shlens, J.; and Szegedy, C.
\newblock 2014.
\newblock Explaining and harnessing adversarial examples.
\newblock {\em arXiv preprint arXiv:1412.6572}.

\bibitem[\protect\citeauthoryear{Gunning}{2017}]{gunning2017explainable}
Gunning, D.
\newblock 2017.
\newblock Explainable artificial intelligence (xai).
\newblock {\em Defense Advanced Research Projects Agency (DARPA), nd Web} 2.

\bibitem[\protect\citeauthoryear{Guo \bgroup et al\mbox.\egroup
  }{2017}]{guo2017calibration}
Guo, C.; Pleiss, G.; Sun, Y.; and Weinberger, K.~Q.
\newblock 2017.
\newblock On calibration of modern neural networks.
\newblock In {\em ICML}.

\bibitem[\protect\citeauthoryear{Hecker, Dai, and
  Van~Gool}{2018}]{hecker2018failure}
Hecker, S.; Dai, D.; and Van~Gool, L.
\newblock 2018.
\newblock Failure prediction for autonomous driving.
\newblock In {\em 2018 IEEE Intelligent Vehicles Symposium (IV)},  1792--1799.
\newblock IEEE.

\bibitem[\protect\citeauthoryear{Hein, Andriushchenko, and
  Bitterwolf}{2019}]{hein2019relu}
Hein, M.; Andriushchenko, M.; and Bitterwolf, J.
\newblock 2019.
\newblock Why relu networks yield high-confidence predictions far away from the
  training data and how to mitigate the problem.
\newblock In {\em CVPR}.

\bibitem[\protect\citeauthoryear{Hendrycks and
  Dietterich}{2019}]{hendrycks2019benchmarking}
Hendrycks, D., and Dietterich, T.
\newblock 2019.
\newblock Benchmarking neural network robustness to common corruptions and
  perturbations.
\newblock {\em ICLR}.

\bibitem[\protect\citeauthoryear{Hendrycks and
  Gimpel}{2016}]{hendrycks2016baseline}
Hendrycks, D., and Gimpel, K.
\newblock 2016.
\newblock A baseline for detecting misclassified and out-of-distribution
  examples in neural networks.
\newblock {\em arXiv preprint arXiv:1610.02136}.

\bibitem[\protect\citeauthoryear{Hendrycks, Lee, and
  Mazeika}{2019}]{hendrycks2019using}
Hendrycks, D.; Lee, K.; and Mazeika, M.
\newblock 2019.
\newblock Using pre-training can improve model robustness and uncertainty.
\newblock {\em arXiv preprint arXiv:1901.09960}.

\bibitem[\protect\citeauthoryear{Hern{\'a}ndez-Orallo \bgroup et
  al\mbox.\egroup }{2019}]{hernandez2019surveying}
Hern{\'a}ndez-Orallo, J.; Mart{\'\i}nez-Plumed, F.; Avin, S.; and
  h{\'E}igeartaigh, S.~{\'O}.
\newblock 2019.
\newblock Surveying safety-relevant ai characteristics.
\newblock In {\em SafeAI@ AAAI}.

\bibitem[\protect\citeauthoryear{Huang \bgroup et al\mbox.\egroup
  }{2018}]{huang2017multi}
Huang, G.; Chen, D.; Li, T.; Wu, F.; van~der Maaten, L.; and Weinberger, K.~Q.
\newblock 2018.
\newblock Multi-scale dense networks for resource efficient image
  classification.
\newblock {\em ICLR}.

\bibitem[\protect\citeauthoryear{Kendall and
  Gal}{2017}]{kendall2017uncertainties}
Kendall, A., and Gal, Y.
\newblock 2017.
\newblock What uncertainties do we need in bayesian deep learning for computer
  vision?
\newblock In {\em NIPS},  5574--5584.

\bibitem[\protect\citeauthoryear{Koopman and
  Wagner}{2017}]{koopman2017autonomous}
Koopman, P., and Wagner, M.
\newblock 2017.
\newblock Autonomous vehicle safety: An interdisciplinary challenge.
\newblock {\em IEEE Intelligent Transportation Systems Magazine} 9(1):90--96.

\bibitem[\protect\citeauthoryear{Lakshminarayanan, Pritzel, and
  Blundell}{2017}]{lakshminarayanan2017simple}
Lakshminarayanan, B.; Pritzel, A.; and Blundell, C.
\newblock 2017.
\newblock Simple and scalable predictive uncertainty estimation using deep
  ensembles.
\newblock In {\em NIPS}.

\bibitem[\protect\citeauthoryear{LeCun, Bengio, and
  Hinton}{2015}]{lecun2015deep}
LeCun, Y.; Bengio, Y.; and Hinton, G.
\newblock 2015.
\newblock Deep learning.
\newblock {\em nature} 521(7553):436--444.

\bibitem[\protect\citeauthoryear{Lee \bgroup et al\mbox.\egroup
  }{2017}]{lee2017training}
Lee, K.; Lee, H.; Lee, K.; and Shin, J.
\newblock 2017.
\newblock Training confidence-calibrated classifiers for detecting
  out-of-distribution samples.
\newblock {\em arXiv preprint arXiv:1711.09325}.

\bibitem[\protect\citeauthoryear{Lee, Eum, and Kwon}{2019}]{lee2019me}
Lee, H.; Eum, S.; and Kwon, H.
\newblock 2019.
\newblock Me r-cnn: Multi-expert r-cnn for object detection.
\newblock {\em IEEE Transactions on Image Processing}.

\bibitem[\protect\citeauthoryear{Leike \bgroup et al\mbox.\egroup
  }{2017}]{leike2017ai}
Leike, J.; Martic, M.; Krakovna, V.; Ortega, P.~A.; Everitt, T.; Lefrancq, A.;
  Orseau, L.; and Legg, S.
\newblock 2017.
\newblock Ai safety gridworlds.
\newblock {\em arXiv preprint arXiv:1711.09883}.

\bibitem[\protect\citeauthoryear{McAllister \bgroup et al\mbox.\egroup
  }{2017}]{mcallister2017concrete}
McAllister, R.; Gal, Y.; Kendall, A.; Van Der~Wilk, M.; Shah, A.; Cipolla, R.;
  and Weller, A.~V.
\newblock 2017.
\newblock Concrete problems for autonomous vehicle safety: Advantages of
  bayesian deep learning.
\newblock IJCAI.

\bibitem[\protect\citeauthoryear{Michelmore, Kwiatkowska, and
  Gal}{2018}]{michelmore2018evaluating}
Michelmore, R.; Kwiatkowska, M.; and Gal, Y.
\newblock 2018.
\newblock Evaluating uncertainty quantification in end-to-end autonomous
  driving control.
\newblock {\em arXiv preprint arXiv:1811.06817}.

\bibitem[\protect\citeauthoryear{Mohseni \bgroup et al\mbox.\egroup
  }{2020}]{mohseni2020ood}
Mohseni, S.; Pitale, M.; Yadawa, J.; and Wang, Z.
\newblock 2020.
\newblock Self-supervised learning for generalizable out-of-distribution
  detection.
\newblock In {\em AAAI}.

\bibitem[\protect\citeauthoryear{Mohseni, Jagadeesh, and
  Wang}{2019}]{mohseni2019predicting}
Mohseni, S.; Jagadeesh, A.; and Wang, Z.
\newblock 2019.
\newblock Predicting model failure using saliency maps in autonomous driving
  systems.
\newblock {\em ICML Workshop on Uncertainty \& Robustness in Deep Learning}.

\bibitem[\protect\citeauthoryear{Ortega and Maini}{2018}]{ortega2018threepilar}
Ortega, P.~A., and Maini, V.
\newblock 2018.
\newblock Building safe artificial intelligence: specification, robustness, and
  assurance.
\newblock
  \url{https://medium.com/deepmindsafetyresearch/building-safe-artificial-intelligence-52f5f75058f1}.

\bibitem[\protect\citeauthoryear{Papernot \bgroup et al\mbox.\egroup
  }{2016}]{papernot2016distillation}
Papernot, N.; McDaniel, P.; Wu, X.; Jha, S.; and Swami, A.
\newblock 2016.
\newblock Distillation as a defense to adversarial perturbations against deep
  neural networks.
\newblock In {\em 2016 IEEE Symposium on Security and Privacy (SP)},  582--597.
\newblock IEEE.

\bibitem[\protect\citeauthoryear{Salay, Queiroz, and
  Czarnecki}{2017}]{salay2017analysis}
Salay, R.; Queiroz, R.; and Czarnecki, K.
\newblock 2017.
\newblock An analysis of iso 26262: Using machine learning safely in automotive
  software.
\newblock {\em arXiv preprint arXiv:1709.02435}.

\bibitem[\protect\citeauthoryear{Seshia \bgroup et al\mbox.\egroup
  }{2018}]{seshia2018formal}
Seshia, S.~A.; Desai, A.; Dreossi, T.; Fremont, D.~J.; Ghosh, S.; Kim, E.;
  Shivakumar, S.; Vazquez-Chanlatte, M.; and Yue, X.
\newblock 2018.
\newblock Formal specification for deep neural networks.
\newblock In {\em International Symposium on Automated Technology for
  Verification and Analysis},  20--34.
\newblock Springer.

\bibitem[\protect\citeauthoryear{Seshia, Sadigh, and
  Sastry}{2016}]{seshia2016towards}
Seshia, S.~A.; Sadigh, D.; and Sastry, S.~S.
\newblock 2016.
\newblock Towards verified artificial intelligence.
\newblock {\em arXiv preprint arXiv:1606.08514}.

\bibitem[\protect\citeauthoryear{Smith and Gal}{2018}]{smith2018understanding}
Smith, L., and Gal, Y.
\newblock 2018.
\newblock Understanding measures of uncertainty for adversarial example
  detection.
\newblock {\em arXiv preprint arXiv:1803.08533}.

\bibitem[\protect\citeauthoryear{Tang \bgroup et al\mbox.\egroup
  }{2019}]{tang2019pamtri}
Tang, Z.; Naphade, M.; Birchfield, S.; Tremblay, J.; Hodge, W.; Kumar, R.;
  Wang, S.; and Yang, X.
\newblock 2019.
\newblock Pamtri: Pose-aware multi-task learning for vehicle re-identification
  using highly randomized synthetic data.
\newblock In {\em ICCV}.

\bibitem[\protect\citeauthoryear{Varshney}{2016}]{varshney2016engineering}
Varshney, K.~R.
\newblock 2016.
\newblock Engineering safety in machine learning.
\newblock In {\em 2016 Information Theory and Applications Workshop (ITA)},
  1--5.
\newblock IEEE.

\bibitem[\protect\citeauthoryear{Vyas \bgroup et al\mbox.\egroup
  }{2018}]{vyas2018out}
Vyas, A.; Jammalamadaka, N.; Zhu, X.; Das, D.; Kaul, B.; and Willke, T.~L.
\newblock 2018.
\newblock Out-of-distribution detection using an ensemble of self supervised
  leave-out classifiers.
\newblock In {\em ECCV},  550--564.

\bibitem[\protect\citeauthoryear{Wang \bgroup et al\mbox.\egroup
  }{2018}]{wang2018efficient}
Wang, S.; Pei, K.; Whitehouse, J.; Yang, J.; and Jana, S.
\newblock 2018.
\newblock Efficient formal safety analysis of neural networks.
\newblock In {\em NIPS},  6367--6377.

\bibitem[\protect\citeauthoryear{Wang, Shrivastava, and
  Gupta}{2017}]{wang2017fast}
Wang, X.; Shrivastava, A.; and Gupta, A.
\newblock 2017.
\newblock A-fast-rcnn: Hard positive generation via adversary for object
  detection.
\newblock In {\em CVPR}.

\bibitem[\protect\citeauthoryear{Wu \bgroup et al\mbox.\egroup
  }{2019}]{wu2019delving}
Wu, Z.; Suresh, K.; Narayanan, P.; Xu, H.; Kwon, H.; and Wang, Z.
\newblock 2019.
\newblock Delving into robust object detection from unmanned aerial vehicles: A
  deep nuisance disentanglement approach.
\newblock In {\em ICCV}.

\bibitem[\protect\citeauthoryear{Zeiler and
  Fergus}{2014}]{zeiler2014visualizing}
Zeiler, M.~D., and Fergus, R.
\newblock 2014.
\newblock Visualizing and understanding convolutional networks.
\newblock In {\em ECCV},  818--833.
\newblock Springer.

\bibitem[\protect\citeauthoryear{Zhang and LeCun}{2017}]{zhang2017universum}
Zhang, X., and LeCun, Y.
\newblock 2017.
\newblock Universum prescription: Regularization using unlabeled data.
\newblock In {\em AAAI}.

\bibitem[\protect\citeauthoryear{Zhang \bgroup et al\mbox.\egroup
  }{2019}]{zhang2019dada}
Zhang, X.; Wang, Z.; Liu, D.; and Ling, Q.
\newblock 2019.
\newblock Dada: Deep adversarial data augmentation for extremely low data
  regime classification.
\newblock In {\em ICASSP}.
\newblock IEEE.

\bibitem[\protect\citeauthoryear{Zhong \bgroup et al\mbox.\egroup
  }{2017}]{zhong2017random}
Zhong, Z.; Zheng, L.; Kang, G.; Li, S.; and Yang, Y.
\newblock 2017.
\newblock Random erasing data augmentation.
\newblock {\em arXiv preprint arXiv:1708.04896}.

\bibitem[\protect\citeauthoryear{Zhou and Tuzel}{2018}]{zhou2018voxelnet}
Zhou, Y., and Tuzel, O.
\newblock 2018.
\newblock Voxelnet: End-to-end learning for point cloud based 3d object
  detection.
\newblock In {\em CVPR}.

\end{thebibliography}

\end{document}